\documentclass[letterpaper, 10 pt, conference]{ieeeconf}  

\IEEEoverridecommandlockouts                              

\usepackage{bm}
\usepackage{times}
\usepackage{multicol}
\usepackage[bookmarks=true]{hyperref}

\usepackage{graphicx} 
\usepackage{amsmath} 
\usepackage{amssymb}  
\usepackage{subfigure}
\usepackage{color}
\usepackage{algorithm, algorithmic}

\usepackage{xargs}

\usepackage{units}

\newcommand{\margincomment}[1]{\marginpar{\raggedright \scriptsize {#1}}}

\iftrue 
    \newcommand{\todo}[1]{{\color{blue}[{\bf TODO:} {#1}]}}

    \newcommand{\trimmed}[1]{\margincomment{\color{magenta}[{#1}]}}
    \newcommand{\baa}[1]{\margincomment{\color{red}[{\bf BAA:} {#1}]}}
    \newcommand{\darpa}[1]{\margincomment{\color{green}[{\bf DARPA:} {#1}]}}
\else
    \newcommand{\todo}[1]{}
    \newcommand{\trimmed}[1]{}
    \newcommand{\baa}[1]{}
    \newcommand{\darpa}[1]{}
\fi

\newcommand{\overparen}[1]%
    {\setbox0\hbox{$#1$} \vbox{\ialign{\hss$##$\hss\cr\ifdim
    \wd0<2em\Large\else\Huge\fi\frown\cr \noalign{\kern-2ex} #1\cr}}}

\newcommand{\bmat}[1]{\left[\!\begin{array}{#1}}
\newcommand{\emat}{\end{array}\!\right]}

\newtheorem{remark}{\bf Remark}
\newcommand{\bre}{\begin{remark} }
\newcommand{\ere}{\end{remark}}

\newcommand{\figref}[1]{Figure~\ref{fig:#1}}

\renewcommand{\eqref}[1]{(\ref{eq:#1})}      

\newcommand{\secref}[1]{Section~\ref{sec:#1}}

\graphicspath{{./}{figures/}}

\title{\LARGE \bf
Long-Duration Autonomy for Small Rotorcraft UAS
including Recharging
}

\author{Christian Brommer$^{1^{*}}$, Danylo Malyuta$^{2^{*}}$, Daniel Hentzen$^{3}$, Roland Brockers$^{3}$
\thanks{$^{1}$Control of Networked Systems Group, \newline Alpen-Adria-Universit{\"a}t Klagenfurt. {\small christian.brommer@ieee.org}}%
\thanks{$^{2}$Autonomous Controls Laboratory, University of Washington.\newline
        {\small danylo@uw.edu}}%
\thanks{$^{3}$Jet Propulsion Laboratory, California Institute of Technology.\newline
        {\small \{daniel.r.hentzen$\vert$brockers\}@jpl.nasa.gov}}%
\thanks{$^{*}$C. Brommer and D. Malyuta contributed equally to this work.}%
}

\begin{document}

\maketitle
\thispagestyle{empty}
\pagestyle{empty}

\begin{abstract}
Many unmanned aerial vehicle surveillance and monitoring applications require observations at precise locations over long periods of time, ideally days or weeks at a time (e.g. ecosystem monitoring), which has been impractical due to limited endurance and the requirement of humans in the loop for operation.
To overcome these limitations, we propose a fully autonomous small rotorcraft UAS that is capable of performing repeated sorties for long-term observation missions without any human intervention.
We address two key technologies that are critical for such a system: full platform autonomy including emergency response to enable mission execution independently from human operators, and the ability of vision-based precision landing on a recharging station for automated energy replenishment.
Experimental results of up to 11 hours of fully autonomous operation in indoor and outdoor environments illustrate the capability of our system.
\end{abstract}

\section{Introduction}
Small rotorcraft unmanned aerial systems (UAS) are ideal assets to deploy sensors and instruments in a 3D environment in applications where data acquisition at precise locations is required. Examples span from environmental monitoring applications in precision agriculture scenarios to surveillance applications of specific targets.
Nevertheless, these applications often demand repeated acquisitions of the same target to capture changes in environmental properties (e.g. plant water usage over the diurnal cycle) or to monitor a specific area of interest. 
While rotorcraft UAS can provide measurements at precise locations, their flight endurance is relatively short (typically \unit[10-20]{min}). 
Additionally, deployment and operation currently requires humans in the loop, making continuous observations over long periods of time (e.g. days or weeks) impractical.
In this paper, we present a technology that solves both issues. 
We propose a system that is capable of fully autonomous repeated mission execution, including automated recharging to replenish its batteries after a flight mission is completed.
We believe that platform autonomy and the ability of precision landing for recharging 
are key technologies for enabling true long-duration missions that allow acquiring continuous observations over days or weeks at a time without human interaction.

%
Our key contribution is the development of a high-level autonomy framework that
is executed on-board the UAS and which implements advanced control for take-off and landing, GPS-inertial state estimation, mission execution, system health monitoring including the execution of emergency behaviors, vision-based precision landing, recharging, and mission data handling.
We demonstrate this technology on an Asctec Pelican UAS that is equipped with an Odroid XU4 and a downward facing camera. As a payload, we deploy a FLIR thermal camera to repeatedly acquire surface temperature of a specific target as an example of a mission data product.
%
Deployment of the system only involves the physical setup of the charging station and the definition of an observation area. Once the system is started, all operations are fully autonomous, with no human in the loop.
In the following chapters, we give an overview of related work in \secref{related_work} and introduce the system concept in \secref{system_overview}.
\secref{autonomy_components} explains all key autonomy components that are implemented on-board the UAS, followed by an overview of our control architecture in \secref{control}, experimental results in \secref{experimental_results} and conclusions in \secref{conclusion_future_work}.


\section{Related Work}
\label{sec:related_work}
\begin{figure}%
\includegraphics[width=\columnwidth]{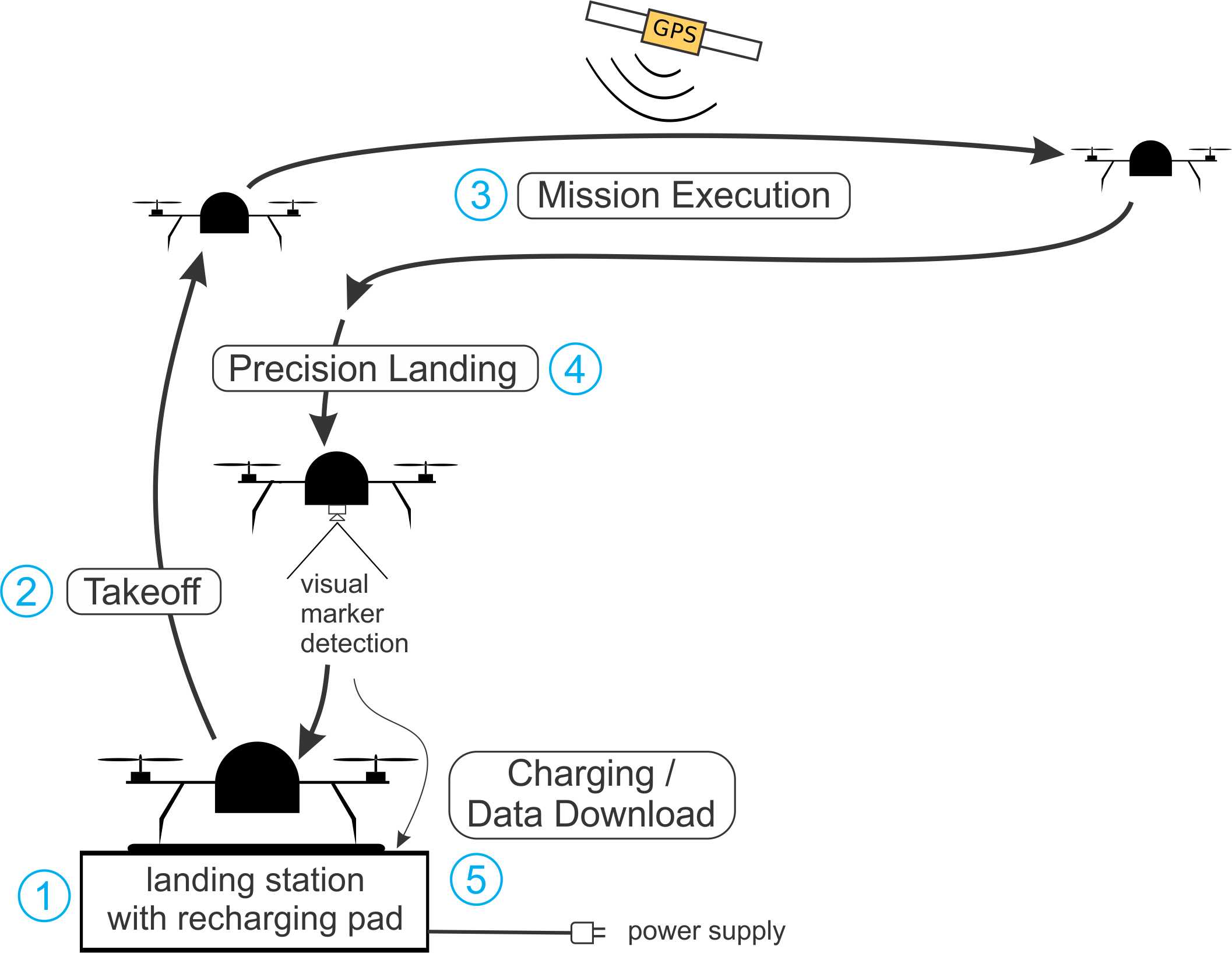}%
\caption{Autonomous UAS data acquisition cycle. 1: the vehicle is placed on the landing station to start the operation. 2: takeoff to safe altitude. 3: mission execution. 4: vision-based precision landing. 5: recharging and data downloading.}%
\label{fig:mission_scheme}%
    \vspace{-4mm}
\end{figure}
Several rotorcraft UAS that are capable of autonomous, long-duration mission execution in constant surveillance scenarios in benign indoor (VICON) environments have been introduced in literature.
Focusing on the recharging solution to extend individual platform flight time and a multi-agent scheme for constant operation, impressive operation times have been demonstrated 
(\cite{Valenti2007MissionHM}: \unit[24]{h} single vehicle experiment; \cite{Mulgaonkar_2014}: \unit[9.5]{h} with multiple vehicles).
Recharging methods vary from wireless charging \cite{Aldhaher_2017} to contact-based charging pads \cite{Mulgaonkar_2014} to battery swap systems \cite{Toksoz_2011}\cite{Suzuki_2012}.
While wireless charging offers the most flexibility since no physical contact has to be made, charging currents are low resulting in excessively long charging times and are hence not an option for quick redeployment. However, interesting results have been shown in \cite{Aldhaher_2017} demonstrating wireless powering of a \unit[35]{g}/\unit[10]{W} micro drone in hover flight.
On the other end of the spectrum, battery swap systems offer immediate redeployment of a UAS, but require sophisticated mechanisms to be able to hot
swap a battery, and a pool of charged batteries that are readily available. 
This makes such systems less attractive for long-term deployment that is maintenance free and cost effective.

In our approach, we use a contact-based charging system which offers a higher charging current and thus shorter charging times than wireless systems, and has less failure modes than a battery swap system. Additionally, the commercial sector is currently developing various contact-based charging solutions for rotorcraft UAS \cite{skysense_2017}\cite{hiveuav_2017}\cite{h3dynamics_2017} that will make charging hardware readily available.
Some of these solutions begin to integrate automated landing capabilities as well, but no complete solution is available today.

GPS-based state estimation for UAS is standard within the commercial sector and used in several research approaches \cite{Kingston2004}\cite{Cheng2014}\cite{Bodo2017}. Most implementations also use magnetometers to estimate yaw with respect to north \cite{Sabatelli2013} or a Gauss Newton Quaternion approach to find rotation between magnetometers and accelerometers \cite{Marins2001}.
Similar to these approaches, we implement a full 3D attitude update from the normalized magnetic field vector, and add a barometer to better estimate altitude \cite{Tanigawa2008}.
While GPS-based state estimation is commonly used for outdoor UAS control, its
accuracy is not sufficient to execute pin-point landing maneuvers on a
reasonably sized recharging platform (i.e. lateral accuracy of several meters
\cite{Center2014}). To address this issue, vision-based landing site detection
using artificial labels has been proposed in several approaches
\cite{Saripalli_2002}\cite{Lins_2015}\cite{Falanga_2017}\cite{Lee_2012}\cite{Yang_2015}\cite{Vetrella_2018}. Similar
to \cite{Chaves_2015}\cite{Vetrella_2018}, we are using multiple markers (AprilTags \cite{wang2016iros}) of different sizes to reliably detect the landing
station from altitude and during approach.


\section{System Overview}
\label{sec:system_overview}
\begin{figure}%
  \centering%
  \includegraphics[width=1.0\columnwidth]{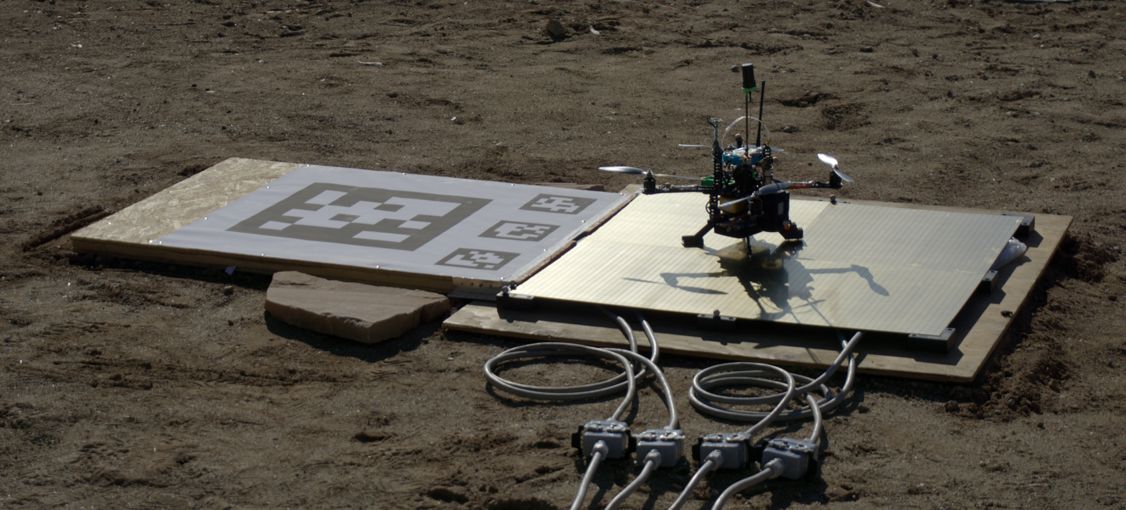}%
  \caption{Landing station with visual bundle, charging pad, and UAS.}%
  \label{fig:landing_pad_with_QR}%
  \vspace{-4mm}
\end{figure}

Our system consists of two major components: the aerial vehicle with its on-board autonomy software (autonomy engine), and a landing station which integrates a 
charging pad, a base station computer for data download,
and a wireless router for communication with the UAS while landed.
%
To deploy the system, a user connects the landing station to a power outlet and
places the UAS on top of the landing surface as shown in \figref{mission_scheme}
and \figref{landing_pad_with_QR}. After entering a waypoint based mission profile, a start command commences the operation of the system. All actions of the UAS from then on are fully autonomous with no human interaction.
The autonomy engine implements various mission specific behaviors to guide the
vehicle, including health monitoring and emergency response.
\figref{mission_scheme} depicts a typical sortie during which behaviors are
executed in the following order.

\begin{figure}[b]%
\includegraphics[width=\columnwidth]{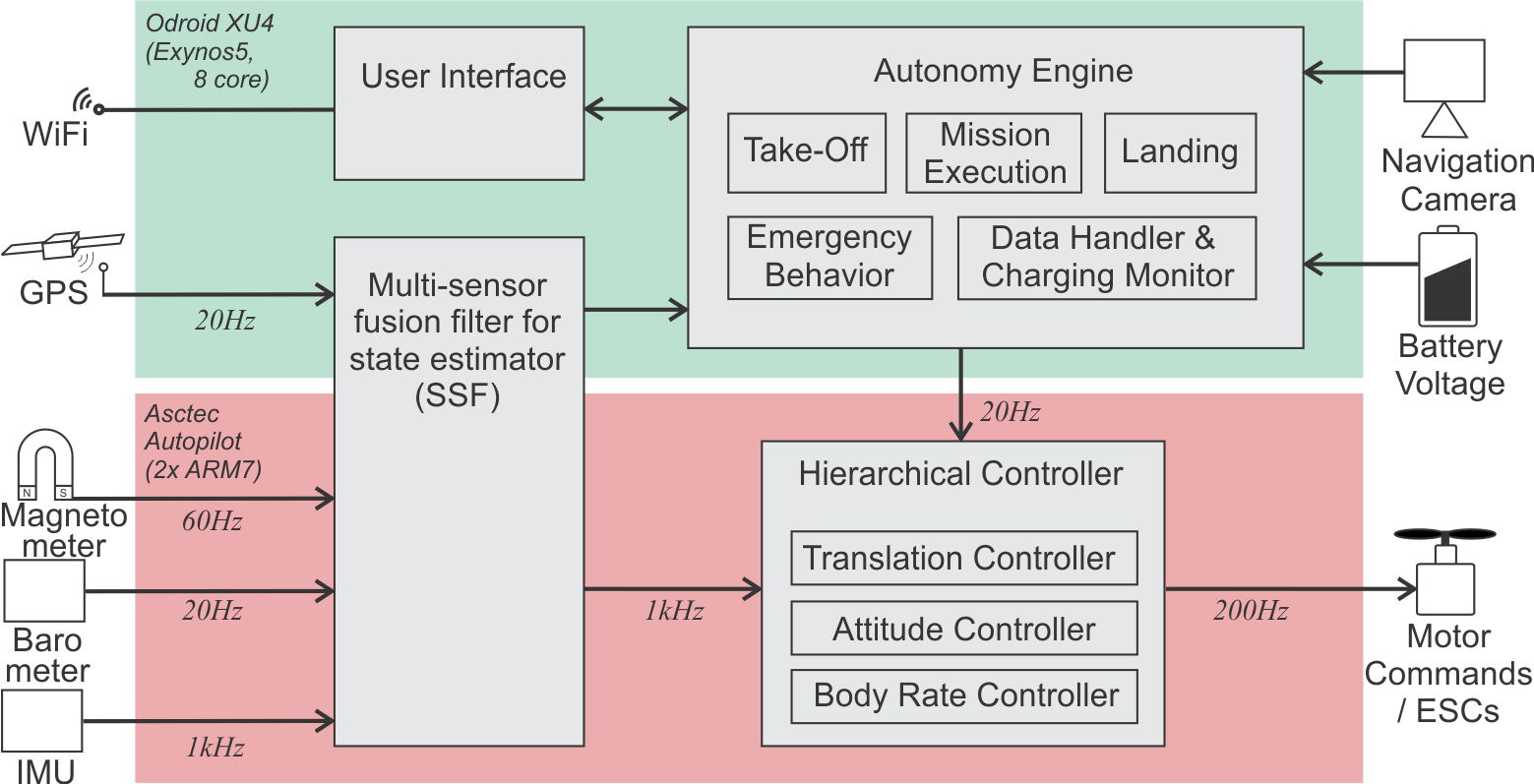}%
\caption{Implementation of autonomy system on-board the UAS. A low level processor performs high rate IMU propagation and executes the control loops, whereas a high-level, more capable embedded computer hosts all computationally expensive algorithms.}%
\label{fig:system_architecture}%
\end{figure}%
Before each take-off, the system initializes the on-board state estimator and passes a series of pre-flight checks that include adequate battery voltage level, and a motor spin-up test. 
The vehicle then performs a take-off maneuver by issuing a vertical velocity command to its low-level velocity controller to climb to an initial altitude.  
Once a safe altitude is reached, the mission execution module within the autonomy engine takes over. 
A mission is defined as a set of waypoints, connected by individual trajectory segments that are calculated using polynomial trajectories following the approach of \cite{Mellinger2011}\cite{Richter2013}.
A trajectory tracking module then plays forward the mission trajectory in time by issuing position references to the low-level position controller.

During mission execution, a system health observer monitors all critical components of the UAS and issues emergency messages to the autonomy engine to implement fail safe behaviors (e.g. return-to-home or emergency-landing on low battery or state estimation failure).

After the mission is completed, the UAS returns to the landing station for
recharging, using the recorded GPS position of the take-off location. Since GPS
accuracy is not sufficient for precision landing, we deploy vision-based landing
site detection that uses April tag fiducials to estimate the accurate position
of the landing station from a safe altitude with a downward looking camera
\cite{wang2016iros}.
%
%
Once the landing location is accurately detected, a landing maneuver is executed by first hovering over the landing location and then issuing a trajectory with a fixed descent velocity that passes through the center of the landing station surface. Touch down is detected by detecting zero velocity below a minimum altitude threshold.

To minimize recharging time, we use a contact based, commercially available charging solution \cite{skysense_2017} to provide ample charging current to the on-board charging electronics. The charging pad consists of an array of individual contact patches that cover a \unit[90]{cm}~x~\unit[90]{cm} flat area. 
Charging is triggered by the charging pad electronics once contact to the charger on-board the UAS is made via leg contacts. 
Charging status is monitored on-line by the autonomy engine, which prepares for the next sortie once batteries are sufficiently charged.
While landed, the vehicle connects to the base station computer via WiFi to
downlink mission data and to receive eventually updated mission plans (e.g. mission-end command from a user).

%
%
%
%

\section{Key Autonomy Components}
\label{sec:autonomy_components}

\figref{system_architecture} depicts our high-level autonomy architecture,
deployed jointly on the Odroid XU4 for computationally intensive tasks and the
Asctec Autopilot for low-level functionality. The following
sections further describe this implementation.
\subsection{GPS-based state estimation}
\label{sec:state_estimation}
\newcommand{\skewmat}[1]{\left[#1\right]_{\!\times}}
To generate high-frequency pose estimates for vehicle control, we deploy a
multi-sensor fusion approach that uses a single Extended Kalman Filter (EKF)
formulation to fuse inertial measurements (\unit[200]{Hz}) with various onboard sensor
measurements such as 3D GPS position (\unit[20]{Hz}), 2D GPS velocity (\unit[20]{Hz}),
barometric pressure (\unit[20]{Hz}) and 3D magnetic field (\unit[60]{Hz}). Our
implementation is based on the SSF framework \cite{Weiss2013} which was extended
to allow the use of multiple sensor inputs for the EKF. 

The EKF state includes vehicle position $p_{wi}$ (at the location of the IMU expressed in the world frame), velocity $v_{wi}$, attitude $q_{wi}$, IMU and pressure sensor biases $b_\omega$, $b_a$, $b_p$, and inter sensor extrinsic calibrations (Eq. \ref{eq:states}).
%
%
\begin{equation}
x=\left[ p_{wi}^T,v_{wi}^T,q_{wi}^T,b_\omega^T,b_a^T,p_{ig}^T,q_{im}^T,m_w^T,b_p^T,p_{ip}^T \right]^T,
\label{eq:states}
\end{equation}

Eq. \ref{eq:state_updates} illustrates the measurement estimates used for the state update (GPS position $\hat{z}_\text{gps\_pos}$, GPS velocity $\hat{z}_\text{gps\_vel}$, pressure height $\hat{z}_\text{press\_height}$ and magnetometer reading $\hat{z}_\text{3D\_mag}$).
%
\begin{subequations}
\begin{align}
\label{eq:gps_pos_update}
\hat{z}_\text{gps\_pos} &= \hat{p}_{wi} + C_{(\hat{q}_{wi})} \ \hat{p}_{ig},\\
\label{eq:gps_vel_update}
\hat{z}_\text{gps\_vel} &= \hat{v}_{wi} + \skewmat{\omega_i} \ \hat{p}_{ig},\\
\label{eq:pressure_height}
\hat{z}_\text{press\_height} &= [0,0,1]^T (\hat{p}_{wi} + C_{(\hat{q}_{wi})} \ \hat{p}_{ig}) + \hat{b}_p,\\
\label{eq:mag_update}
\hat{z}_\text{3D\_mag} &= C_{(\hat{q}_{im})}^T \ C_{({\hat{q}_{wi}})}^T \ \hat{m}_w.
\end{align}
\label{eq:state_updates}
\end{subequations}%
Note, that in Eq. \ref{eq:state_updates} rotation matrices (e.g. $C_{(\hat{q}_{wi})}$) are derived from the respective quaternion estimates (e.g. $\hat{q}_{wi}$).
The core states (position, velocity and attitude) are propagated at \unit[1]{kHz} on the
Asctec high level (HL) processor, while the state updates   are performed on the
Odroid~XU4 at the individual sensor rates.

The propagated state is sent from the HL processor to the Odroid~XU4 
at \unit[200]{Hz} for calculation of the state updates. State corrections are then sent from the Odroid XU4 to the HL processor at the GPS
measurement rate (\unit[20]{Hz}). This reduces the bandwidth used by state estimation to communicate with the HL processor which is shared with other system components.

Different to most implementations in the literature, which use the magnetometer
for yaw estimation only \cite{Sabatelli2013}, we use a normalized 3D magnetic
field vector $\hat{m}_w$ to update the full attitude quaternion as shown in
Eq. \ref{eq:mag_update}. This leads to a fully observable pose and improves our
attitude knowledge.

\subsubsection*{Filter initialization}
The EKF is reset before each take-off, the position is initialized
based on current GPS and pressure measurements, and Gyro biases are set to zero,
whereas accelerometer biases are calculated by averaging the latest
accelerometer measurements for \unit[10]{seconds} before the vehicle lifts off. Pressure sensor bias is initialized 
to align current pressure height with GPS height in order to match the world frame height. 
To initialize the attitude of the vehicle before takeoff, we average magnetometer measurements for a yaw estimate and use averaged accelerometer measurements to estimate the gravity vector \cite{Valenti2015}. 
Inter sensor calibrations such as GPS sensor position and magnetometer orientation are initialized once at the beginning of the deployment with manually estimated values. Additionally, the state covariance is initialized based on heuristic values, while cross-correlations of the states are set to zero.

\subsection{Vision-based Precision Landing}
\label{precision_landing}
To estimate the target landing location on the charging pad, the landing station is equipped with a set of visual markers, which are detected from altitude during approach for landing.
To increase detection rate and accuracy from various altitudes, we designed a
visual target that integrates April tag markers \cite{wang2016iros} of different sizes on a flat surface (\figref{landing_pad_with_QR}).
To relate the position of individual markers to the desired landing location, the tag bundle is calibrated once, where the pose of each marker (position and yaw) with respect to a landing point (defined by a separated master tag during calibration) is recorded.
During landing we use rectified images from a downward facing camera in order to
detect the individual tags. Then, a perspective-$n$-point algorithm estimates
the camera pose in the bundle frame using Levenberg-Marquardt reprojection error
minimization. Individual estimates are processed by a Recursive Least Squares
(RLS) filter, with exponential forgetting, which outputs a more accurate and
precise position and yaw estimate of the landing pad (we assume zero roll and
pitch), even when the tag bundle temporarily moves out of sight of the
downward facing camera. The RLS filter output is used by the landing autopilot for
pinpoint landing, as explained in \secref{state_machine}. Our use of RLS to
achieve a smoother and more accurate landing site estimate goes beyond
\cite{Vetrella_2018}, where raw tag detections are used.
%
%
%

\subsection{High-level Autonomy Engine}
\label{sec:state_machine}
Execution of the full-cycle autonomous data acquisition mission shown in
\figref{mission_scheme} is enabled by an autonomy engine that implements mission
logic using a set of Finite State Machines (FSM).

\figref{mission_scheme} illustrates a high-level overview of the nominal
long-term data acquisition cycle.
The state machines for each phase of a flight -- takeoff, mission execution, and landing -- are stand-alone subsystems that execute the logic associated with each phase based on an external call to action. These phase-specific state machines are called \textit{autopilots}. 
An additional over-arching logic is needed to issue transitions between individual phases in an appropriate sequence. This is achieved via a high-level \textit{master} state machine. 
The master can wake and, if necessary, abort the autopilots via ROS service or action calls. All state machines are executed in parallel threads at the guidance frequency (\unit[20]{Hz}).


\subsubsection{Master State Machine}
\label{master_sm}
\begin{figure}%
  \centering%
  \includegraphics[width=0.7\columnwidth]{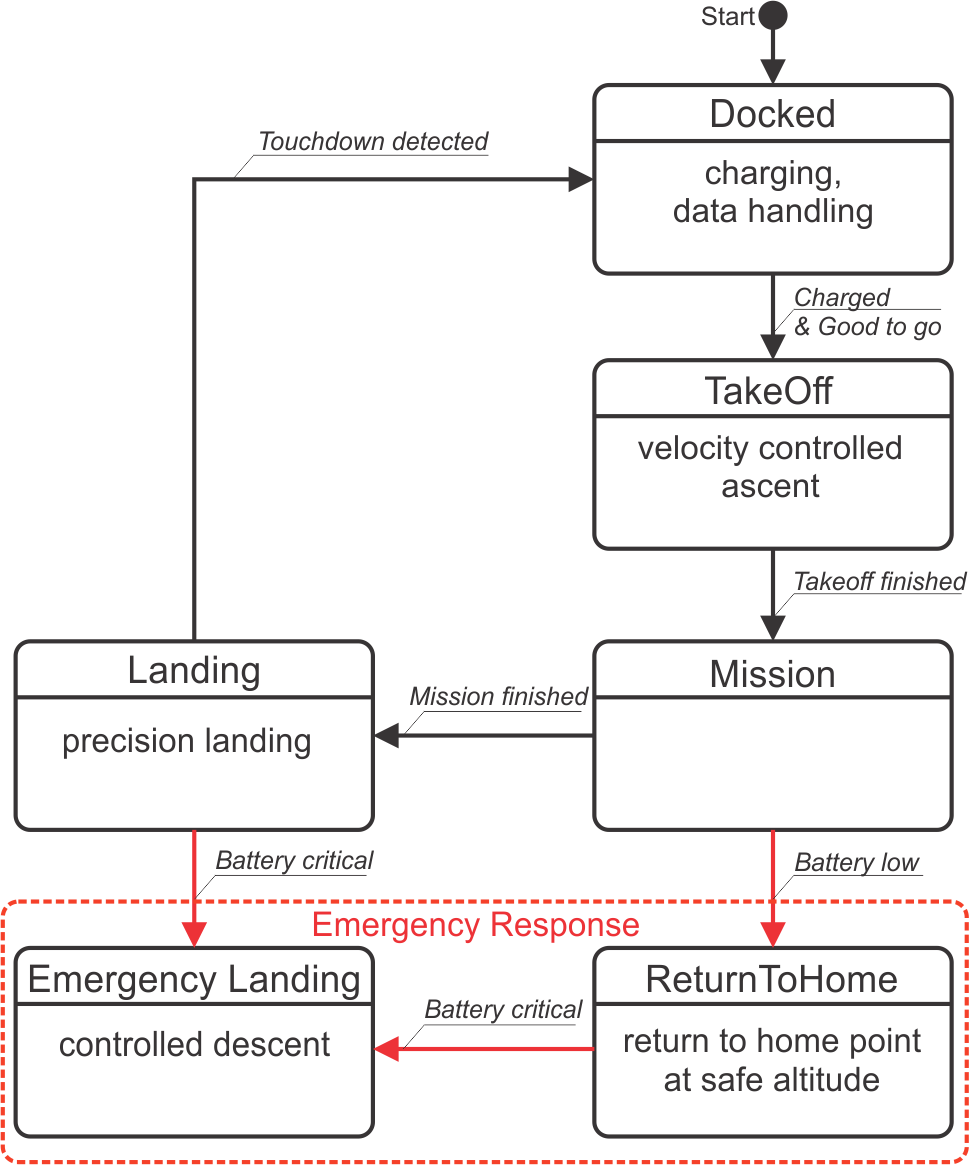}%
  \caption{Master state machine.}%
  \label{fig:master_state_machine}%
\end{figure}

The master state machine coordinates calls to individual autopilot FSMs. In
contrast to the computation-heavy autopilots, the master performs no computations itself. 
Its sole responsibility is to initiate event-based state transitions.
The master state machine is illustrated in \figref{master_state_machine}. 
State transitions are executed according to the mission
cycle of \figref{mission_scheme}. Additional user input can force certain transitions for testing purposes
(\textsc{ForceTakeoff}, \textsc{ForceLand},  \textsc{ReturnToHome} and \textsc{EmergencyLand}).

%

\subsubsection{Takeoff Autopilot}
\label{takeoff_sm}
After successfully validating motor nominal performance, reinitializing SSF
after the prolonged charging phase, and memorizing the current horizontal
location in a local SQLite database, the takeoff autopilot commands a
velocity-controlled takeoff, tracking a vertical velocity trajectory that takes
it from the initial position on the charging pad to a safe altitude.


\subsubsection{Mission Autopilot}
\label{mission_sm}
The mission autopilot is responsible for executing the actual data acquisition
 mission as defined by the user in the form of individual waypoints and
hover times.
The waypoint list gets converted into a polynomial trajectory, using
differential flatness concepts presented in \cite{Richter2013}\cite{Burri2015}. The mission trajectory consists of a smooth sequence of fully
constrained polynomial segments, thus no on-board optimization is required and
the resulting trajectory has a deterministic and well-behaved shape at all
times. The computed trajectory is sampled at the guidance frequency and is
passed as a reference to the position controller described in
Section~\ref{position_controller}.  For emergency response, a
\textsc{ReturnToHome} event is defined. It can be invoked internally (triggered
by a battery voltage below a pre-defined threshold) or externally by the user
(via a ROS service), with the effect of aborting the mission and returning the
UAS to the charging pad.  A mission can be defined as \textsc{single} where the
vehicle returns for landing after the mission is completed, or
\textsc{continuous}, in which case the mission restarts after the last waypoint
has been visited, and the mission end is autonomously triggered by the depletion of
the battery.


\subsubsection{Landing Autopilot}
\label{landing_sm}
Once the mission is finished, the vehicle returns to the previously recorded
take-off position at a safe approach altitude.  At this point, the RLS
estimator is initialized while the vehicle is hovering to start the detection of
the landing location. The estimated yaw is used to align the vehicle attitude
such that the camera points towards the visual markers, while the landing location estimate is used to align the vehicle's lateral position with the desired landing location (i.e. the center of the charging pad). The vehicle then performs a constant velocity descent
until touchdown is detected based on the estimated distance between the vehicle and the charging pad and a threshold on the descent velocity.


\subsubsection{Emergency Landing Autopilot}
\label{emergency_landing_sm}
Emergency landing is triggered as an emergency response to a critically low
battery event. In this case the vehicle will descend at the current position
with a predefined vertical velocity until touchdown is detected based on a
near-zero velocity threshold. This behavior has been successfully tested in
software-in-the-loop simulation and in indoor flights.


\section{Control Architecture}
\label{sec:control}
\begin{figure}[b]
  \centering
  \includegraphics[width=1\columnwidth]{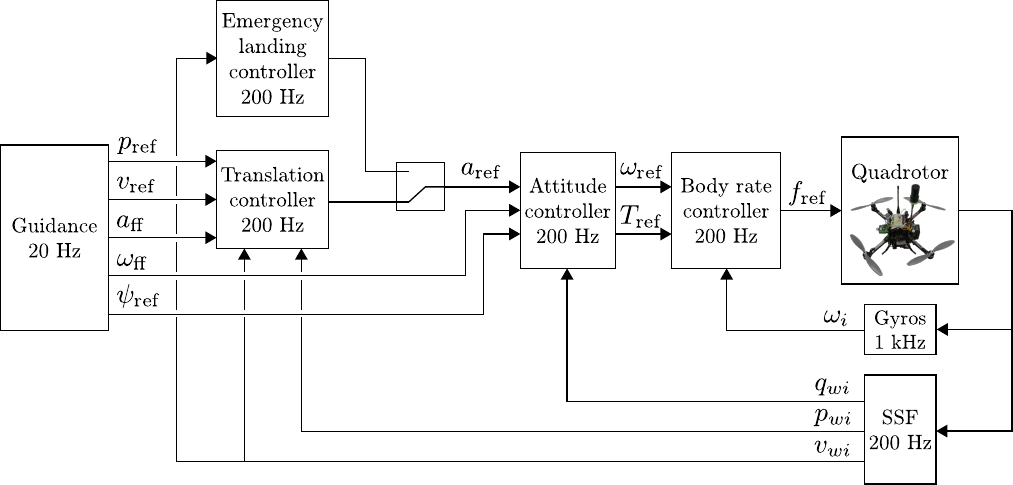}
  \caption{Three-stage cascaded control architecture.}
  \label{fig:control_diagram}
\end{figure}
We use a three-stage cascaded control architecture shown in
Figure~\ref{fig:control_diagram} to maneuver the UAS along the reference
trajectories computed by the autopilots. The stages are, from outer- to
inner-most loop: translation, attitude and body rate controllers. A cascaded
controller has the advantage that inner loops provide faster disturbance
rejection and reduce the effect of nonlinearities \cite{Skogestad2005}, which is
beneficial for accurate tracking.
The controller is executed on the Asctec HL processor.

\subsection{Outer Loop}
\label{position_controller}
A trajectory segment is defined by a reference position $p_{\text{ref}}$, a reference velocity $v_{\text{ref}}$, a
reference yaw $\psi_{\text{ref}}$, an optional feed-forward acceleration $a_{\text{ff}}$ and an optional body rate $\omega_{\text{ff}}$. 
All quantities are obtainable from our polynomial trajectories via differential flatness theory \cite{Mellinger2011}.
The outer control loop which consists of a translation and an attitude controller is responsible for converting this reference
trajectory into a reference for the body rate controller.

\subsubsection{Translation Control}
The position controller consists of a PID controller with a
pre-filtered reference, which outputs a desired vehicle acceleration
$a_{\text{ref}}$. Since $a_{\text{ref}}$ defines a reference thrust vector, the
translation controller may be thought of as a thrust vector calculator. 

\subsubsection{Attitude Control}
The attitude controller tracks $a_{\text{ref}}$ and $\psi_{\text{ref}}$ by
generating a body rate reference $\omega_{\text{ref}}$ and a collective thrust
reference $T_{\text{ref}}$ for the body rate controller.  Our implementation of
the attitude controller is based on the globally asymptotically stable,
quaternion-based controllers introduced in \cite{Brescianini2013} and
\cite{Faessler2015}.
\subsection{Inner Loop}
\label{attitude_controller}
The body rate controller forms the inner-most loop of the cascaded flight
control system. It tracks $\omega_{\text{ref}}$ by computing reference body
torques, following the feedback-linearizing control scheme presented in
\cite{Faessler2016}. These along with $T_{\text{ref}}$ are then converted into
individual reference motor thrusts $f_{\text{ref}}$ that are mapped to propeller
speeds via a motor calibration map. As in \cite{Faessler2016}, our controller
uses a thrust saturation scheme which prioritizes roll and pitch torques (the
most important ones for stability).


\section{Experimental Results}
\label{sec:experimental_results}
\begin{figure}[t]%
\includegraphics[width=\columnwidth]{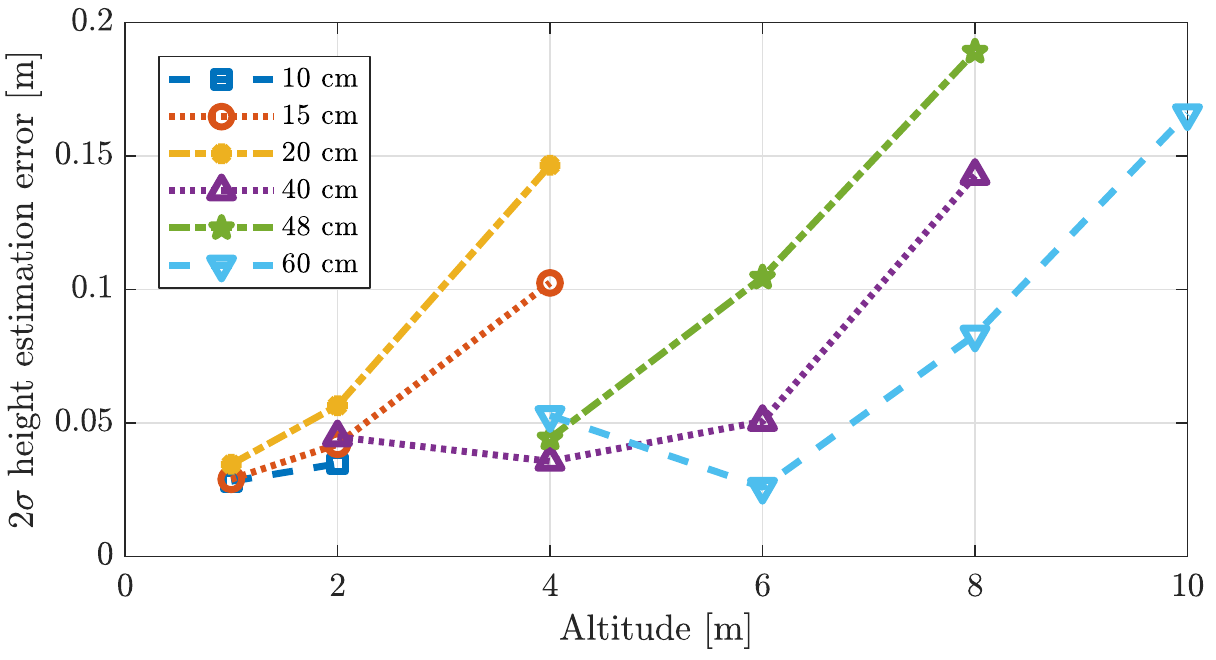}%
\caption{Distance estimation error of AprilTag2 detector for various tag sizes and heights.}%
\label{fig:april_tag_height_error}%
\end{figure}
To evaluate and verify the UAS autonomy framework, we performed a series of indoor and outdoor experiments. 
For indoor experiments, we are using VICON measurements as ground truth and as position inputs to our state estimation filter. 
For outdoor experiments, the system deploys the GPS-based state estimator described in \secref{state_estimation}. For validation purposes , we are additionally using RTK-GPS measurements for ground truthing in outdoor experiments.

\subsection{Precision Landing Performance}
To design a visual target to label the charging station, we first analyzed detection accuracy of the the AprilTag2 algorithm for single tags at
various observation distances using the visual camera that is deployed on our UAS (MatrixVision blueFOX-200wG with a 100$^{\circ}$ FOV lens) in a stationary configuration. 
%
%
\begin{figure}[t]%
	\centering
	\includegraphics[width=0.95\columnwidth]{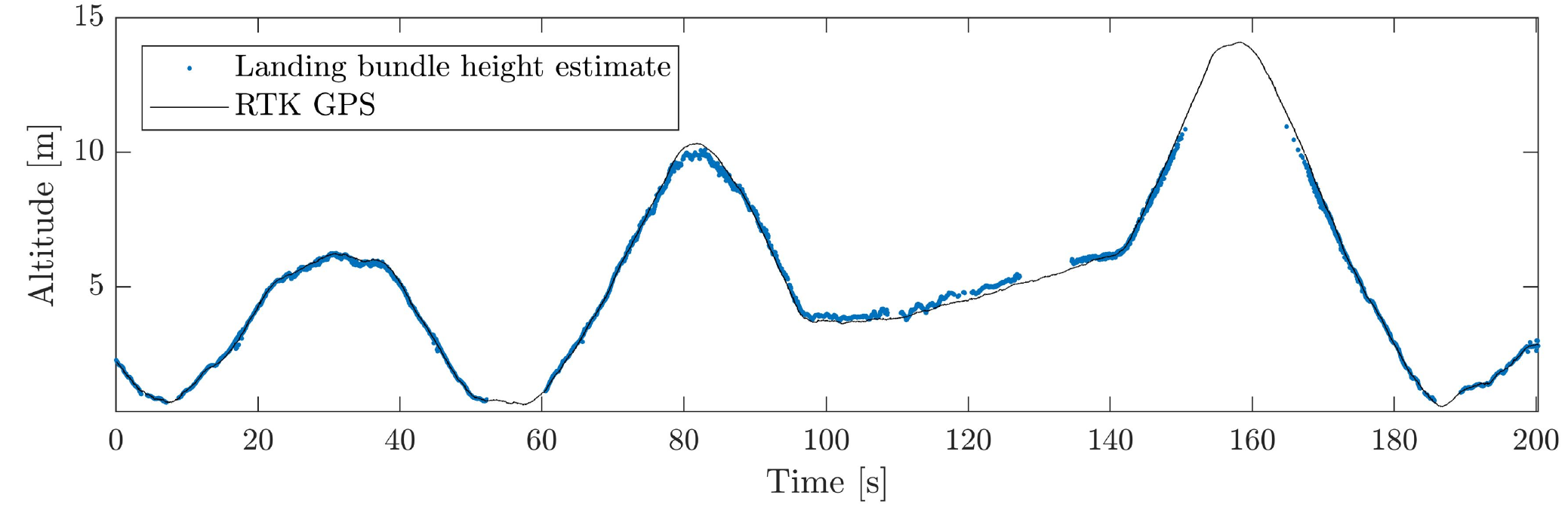}%
	\caption{Comparison of landing bundle height estimate with ground truth altitude during a test flight of the UAS above the landing station. Ground truth is calculated as the difference between the RTK-GPS altitude during flight and when landed on the charging station.}%
	\label{fig:landing_target_height_accuracy}%
\end{figure}
\begin{figure}[t]%
	\centering
	\includegraphics[width=0.95\columnwidth]{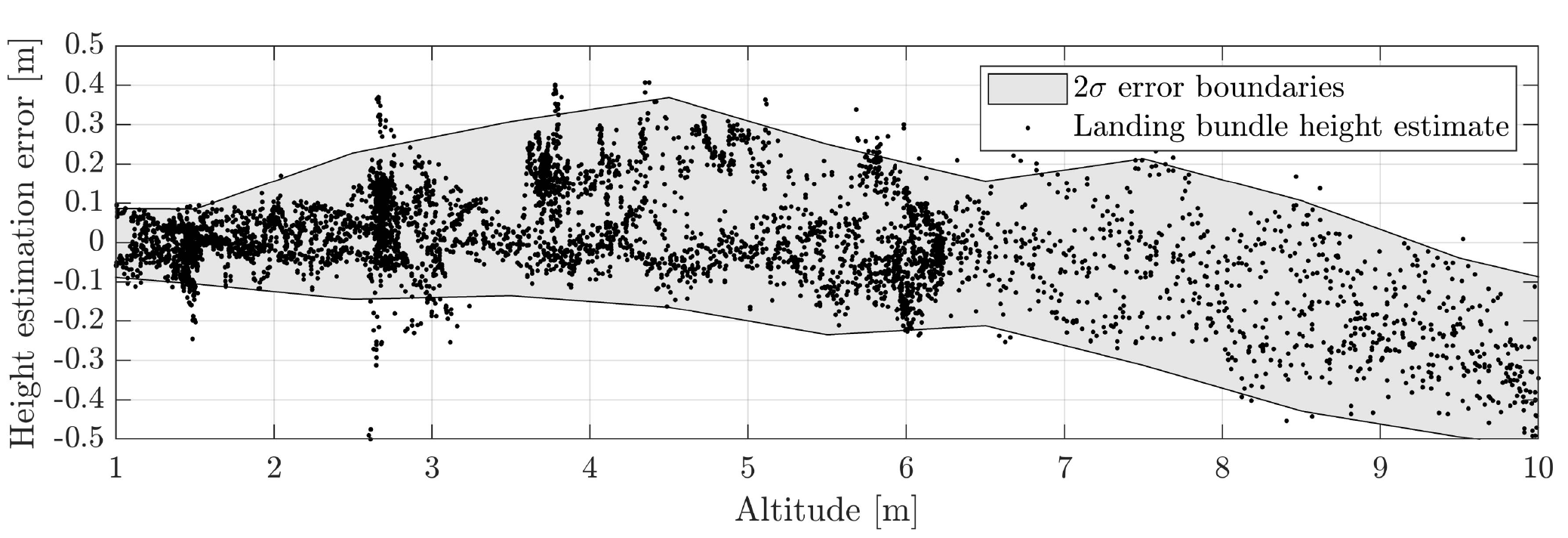}%
	\caption{Landing bundle height error as a function of flight altitude.}%
	\label{fig:landing_target_height_accuracy_error}%
\end{figure}
\begin{figure}[b]%
\includegraphics[width=0.80\columnwidth]{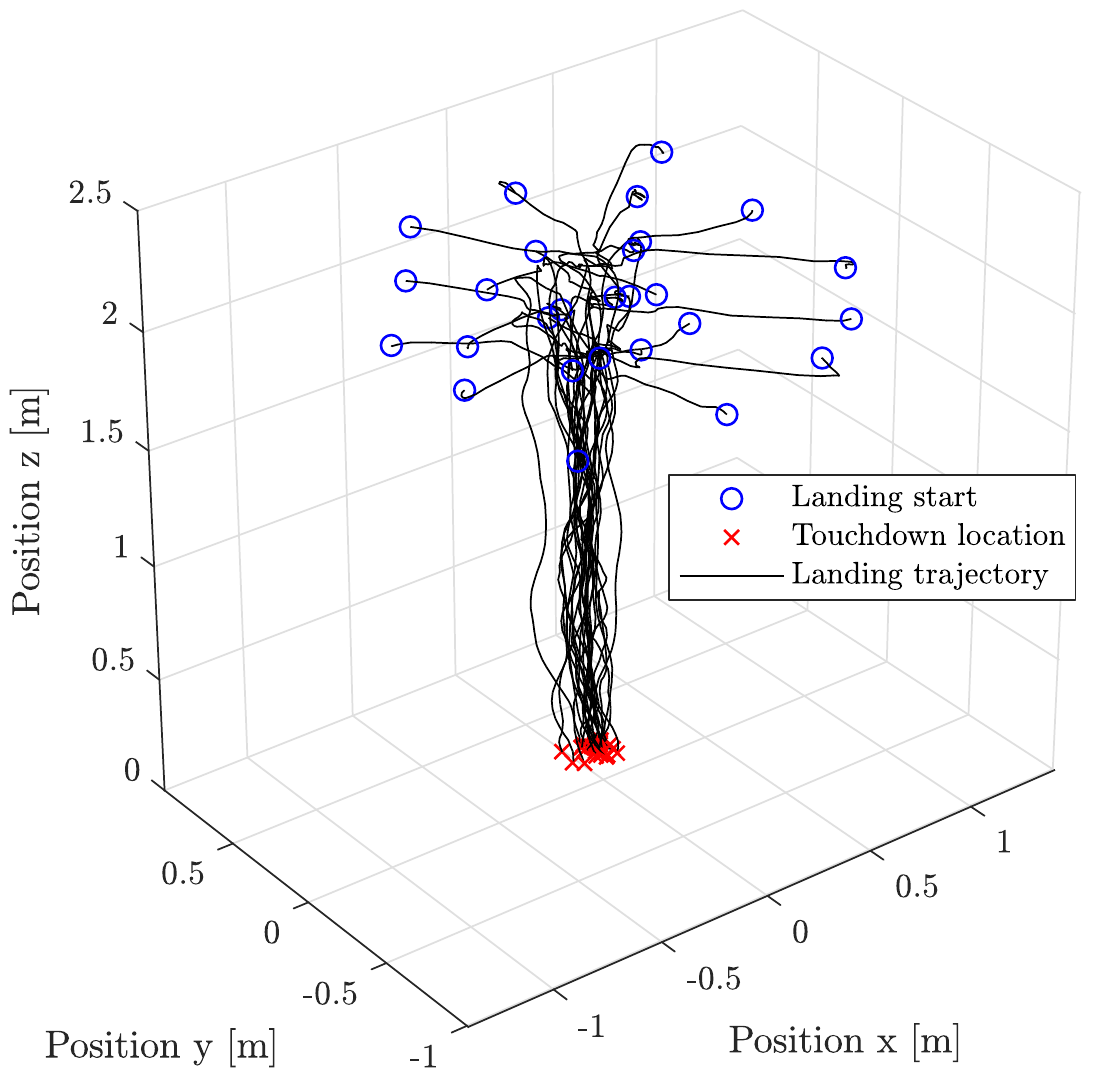}%
\centering
\caption{Landing trajectories for approaches from a set of different waypoints in indoor environment.}%
\label{fig:landing_precision_trajectories}%
\end{figure}
\figref{april_tag_height_error} shows the standard deviation of the AprilTag2 height estimation for various tag sizes. 
For a given tag size, the estimation accuracy is decreasing with increasing distance. 
Large tag sizes (\unit[48]{cm}, \unit[60]{cm}) are not detected at distances below \unit[2]{m}, 
while the small tag sizes (\unit[10]{cm}, \unit[15]{cm}, \unit[20]{cm}) cease to be detected above \unit[4]{m}. 
For an initial approach height of \unit[4]{m}, we selected a \unit[48]{cm} tag for an expected $2\sigma$ height estimation error of \unit[4.5]{cm} and bundled it with three \unit[15]{cm} tags for increased precision in the final stage of the landing.

\figref{landing_target_height_accuracy_error} illustrates the accuracy of the height estimation above our intended landing point using the bundle height estimator with our visual target during an actual descent of our UAS (\figref{landing_target_height_accuracy}).
At the approach height of \unit[4]{m} the 2$\sigma$ height detection error is \unit[0.32]{m}. The error decreases to below \unit[0.1]{m} in the final stage of the descent. We note that the estimation error variance is larger for the tag bundle in \figref{landing_target_height_accuracy_error} than for the individual tags in \figref{april_tag_height_error}. The increased error is most likely caused by pixel quantization effects due to the motion of the vehicle and a lever arm effect that amplifies position errors at the desired landing location (the charging pad center) as a result of angular errors in the April tag detection.

\begin{figure}%
\includegraphics[width=\columnwidth]{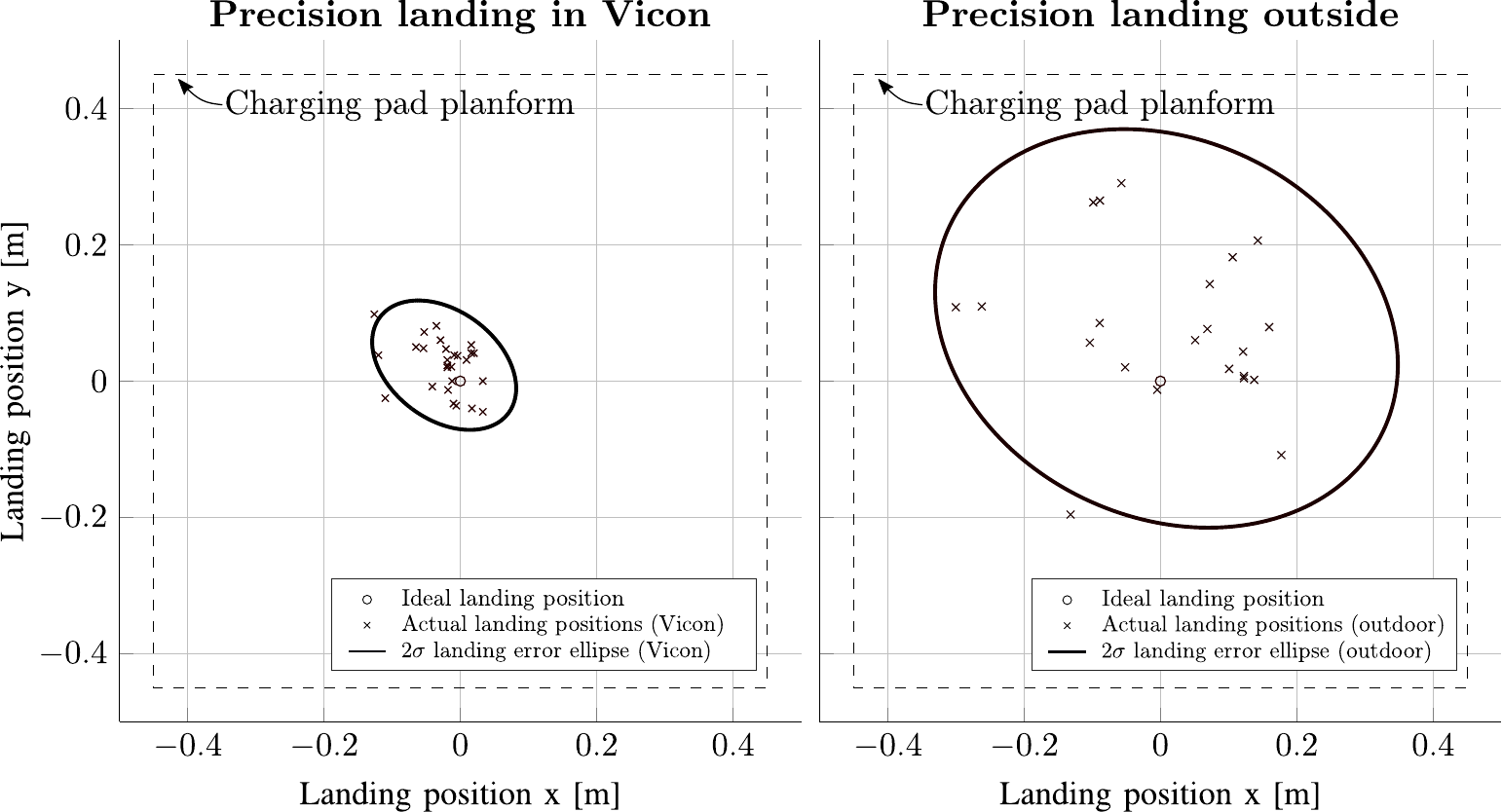}%
\caption{Accuracy of touch-down position in indoor experiment (left), and in outdoor experiment with light wind (right). Ground truth is provided indoors by VICON and outdoors by RTK-GPS.}
\label{fig:landing_ellipse_twins}%
\end{figure}

\figref{landing_precision_trajectories} illustrates the accuracy of autonomous landing with the complete system during an indoor experiment using a VICON system for position estimation and ground truthing. Here, the UAS performed 25 landings in near-optimal conditions. The resulting $2\sigma$ landing error ellipse (\figref{landing_ellipse_twins} (left)) aggregates all detection and control errors. The lateral $2\sigma$ error is \unit[0.11]{m} (major half-axis) and the longitudinal $2\sigma$ error is \unit[0.08]{m} (minor half-axis). \figref{landing_ellipse_twins} (right) repeats this evaluation for outdoor landing tests using RTK-GPS for ground truth and verifies a lateral $2\sigma$ error of \unit[0.37]{m} and a longitudinal $2\sigma$ error of \unit[0.28]{m}, which is well within the margin of the $0.9~\times~0.9$~m recharging platform. 

\begin{figure}[t]%
	\includegraphics[width=\columnwidth]{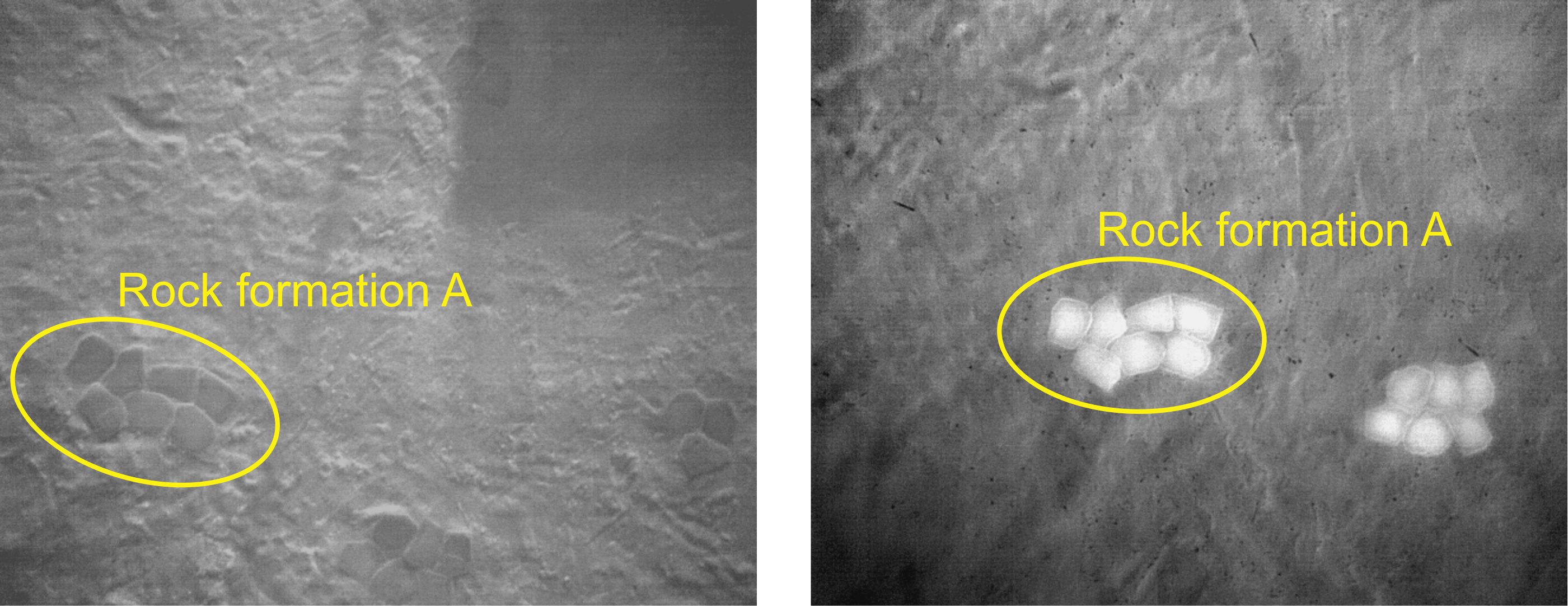}%
	\caption{Thermal images as example mission data product. Brighter pixels show higher temperatures. Left: Cool rock formation in the morning. Right: Rocks heat up during the day.}%
	\label{fig:FLIR_image}%
\end{figure}

\subsection{Long-duration Autonomy Experiments}
\begin{figure*}%
\includegraphics[width=1.0\textwidth]{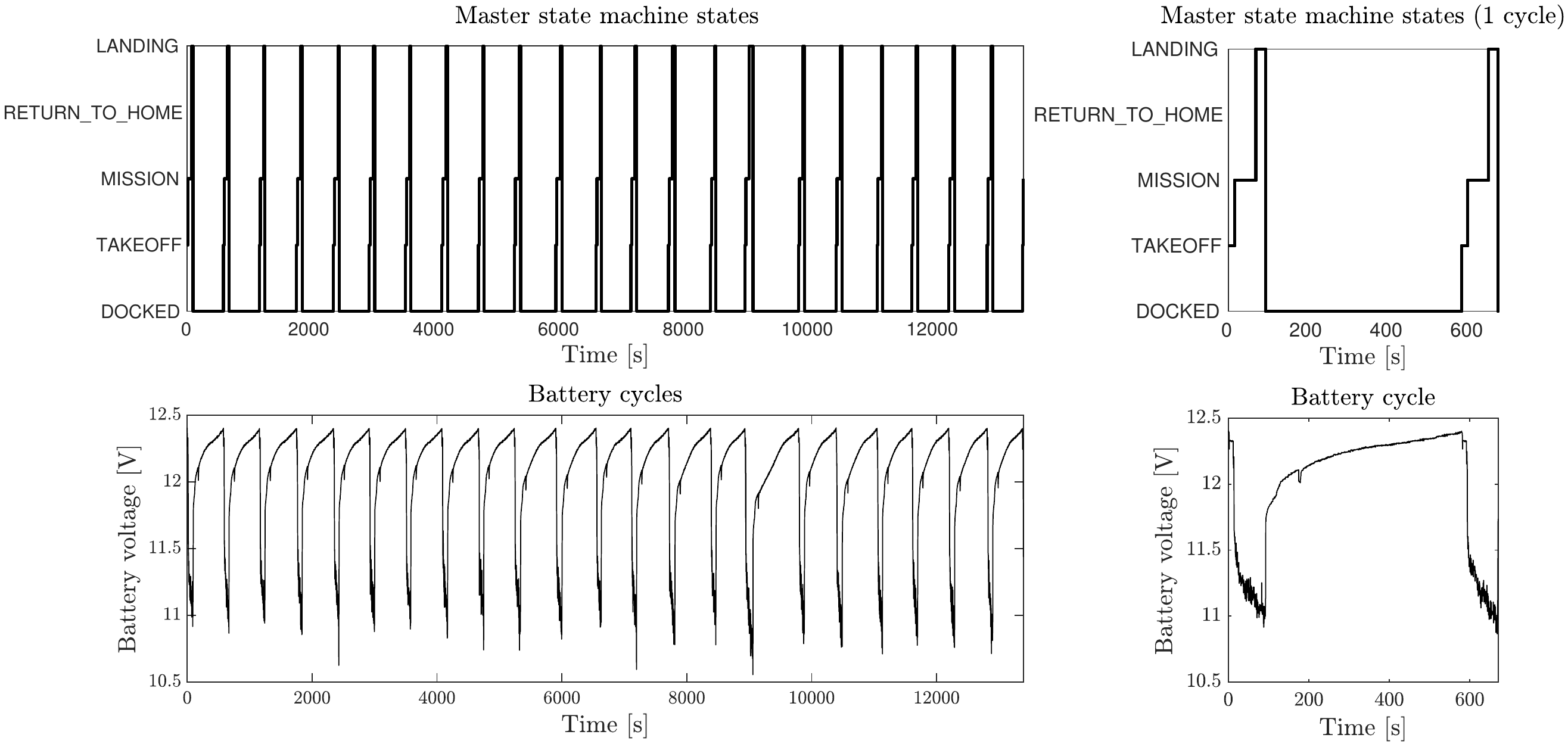}%
\caption{State transitions of the master state machine (top) and battery voltage
  level (bottom) during the \unit[4]{h} outdoor experiment. Left: full experiment. Right: one sortie.}%
\label{fig:longterm_experiment}%
\end{figure*}

We tested the system during a set of indoor and outdoor missions. For component verification, we initially performed indoor experiments in VICON to verify recharging (\unit[11]{h} experiment with 16~sorties). The full system was tested indoors (\unit[10.6]{h} with 48~sorties), and finally outdoors under light wind conditions (\unit[4]{h}, 24~sorties) where the vehicle was commanded to execute a mission profile to visit specific GPS waypoints to capture images with an on-board thermal camera to monitor surface temperatures during the course of the experiment (see \figref{FLIR_image}).
\figref{longterm_experiment} illustrates the state transitions of the master state machine during the \unit[4]{h} outdoor experiment, and the battery voltage reflecting the individual charging cycles of this experiment. 


\section{Conclusions and Future Work}
\label{sec:conclusion_future_work}
We introduced a fully autonomous small rotorcraft UAS that is capable of executing
long-duration data acquisition mission without human intervention. Such a system
enables applications that require continuous long-term observations that are impractical to perform with current UAS. We plan to use our system to deploy a thermal sensor in a precision agriculture scenario to measure land surface temperature (LST) which can be used to estimate water usage over crop fields.

\section*{Acknowledgment}
This research was carried out at the Jet Propulsion Laboratory, California
Institute of Technology, under a contract with the National Aeronautics and
Space Administration. 
The authors would like to thank Stephan Weiss for his inputs on the state estimation approach.

\bibliographystyle{IEEEtran}
\bibliography{combined_ref_no_pages}
\end{document}